\documentclass[pdflatex,sn-mathphys-num]{sn-jnl}

\usepackage{natbib} 
\usepackage{float}
\usepackage{graphicx}%
\usepackage{multirow}%
\usepackage{amsmath,amssymb,amsfonts}%
\usepackage{amsthm}%
\usepackage{mathrsfs}%
\usepackage[title]{appendix}%
\usepackage{xcolor}%
\usepackage{textcomp}%
\usepackage{manyfoot}%
\usepackage{booktabs}%
\usepackage{algorithm}%
\usepackage{algorithmicx}%
\usepackage{algpseudocode}%
\usepackage{listings}%
\usepackage{subcaption}
\usepackage{url}%
\usepackage{hyperref}%
\usepackage{fancyhdr}
\hypersetup{
	colorlinks=true,
	linkcolor=blue,
	citecolor=blue,
	urlcolor=blue
}

\begin{document}
	\fancyhf{} 
	\fancyhead[C]{\small Commments: Presented at \textit{International Conference on Business and Digital Technology}, Bahrain, Springer Nature, 27 November 2025.}
	\renewcommand{\headrulewidth}{0pt} 
	
	\title[Agentic AI Framework for People with Disabilities and Neurodivergence]{Agentic AI Framework for Individuals with Disabilities and Neurodivergence: A Multi-Agent System for Healthy Eating, Daily Routines, and Inclusive Well-Being}
	
	\author*[1]{\fnm{Salman} \sur{Jan}}\email{salman.jan@aou.org.bh}
	
	\author*[2]{\fnm{Toqeer Ali} \sur{Syed}}\email{toqeer@iu.edu.sa}
	
	\author[2]{\fnm{Ali} \sur{Akarma}}\email{443059463@stu.iu.edu.sa}
	
	\author[2]{\fnm{Ahmad} \sur{Ali}}\email{443060103@stu.iu.edu.sa}
	
	\author*[1]{\fnm{Mohammad Riyaz} \sur{Belgaum}}\email{mohammad.riyaz@aou.org.bh}
	
	\affil*[1]{\orgdiv{Faculty of Computer Studies}, \orgname{Arab Open University}, \country{Bahrain}}
	
	\affil*[2]{\orgdiv{Faculty of Computer and Information System}, \orgname{Islamic University of Madinah}, \country{Saudi Arabia}}

	\abstract{
		 	The presented paper suggests a detailed Agentic Artificial Intelligence (AI) model that would enable people with disabilities and neurodivergence to lead a healthier lifestyle and have a more regular day. The system will use a multi-layer structure; it will include an Application and Interface Layer, an Agents Layer, and a Data Source Layer to provide adaptive, transparent, and inclusive support. Fundamentally, a hybrid reasoning engine will synchronize four special-purpose agents, which include: a personalized-nutrition-based, called a Meal Planner Agent; an adaptive-scheduling-based, called a Reminder Agent; an  interactive-assistance-during-grocery-shopping-and-cooking, called a Food Guidance Agent; and a continuous-intake-and-physiological-tracking, called a Monitoring Agent. All the agents interact through a central communicative system called the Blackboard/Event Bus, which allows autonomous interaction and real-time feedback loops with multimedia user interfaces.
		 	
		 	 Privacy-sensitive data sources, including electronic health records (EHRs), nutritional databases, wearable sensors and smart kitchen internet of things, are also included in the framework and placed into a policy-controlled layer, which ensures data safety and compliance with consent.
		 	 Collaborative care and clinician dashboards allow common supervision, discussable artificial intelligence (XAI) modules give brief explanations why a decision was made, making users responsible and reliant.The proposed agentic AI framework is an extension beyond traditional assistive systems since it incorporates inclusiveness, personalization, and accessibility at all levels. It displays the intersection of multi-agent reasoning, multi-modal interfaces, and human-centered design that will enable the development of autonomy, health and digital equity among people with disabilities and neurodivergence.The outcomes of implementation indicate that it can be used to achieve greater adherence, lessen the dependency between caregivers and patients, and empower proactive and personalized health care.
	}
	
	
	\maketitle
	
	\section{Introduction}\label{sec:intro}
	
	The dietary habits and routine are vital to physical and psychological health. Nevertheless, disabled people or neurodivergent people are unique because of the mobility issues, sensory sensibility, and cognitive distortions. Currently, available digital wellness solutions do not provide the desired amount of personalization and adaptive intelligence to overcome such issues~\cite{WHO2020,UNICEF2021}.
	
	The population with disabilities is more than one billion, approximately 15 per cent of the worldwide population~\cite{WHO2020}. Such groups as autism spectrum disorder (ASD), ADHD, or dyslexia are often exposed to obstacles to healthcare and lifestyle assistance~\cite{Kapp2020,SonugaBarke2021}. The existing tools only offer general information and seldom combine cognitive, behavioral and physiological levels that these groups need.
	
	The development of AI has facilitated more integrative interventions. Condition-specific meal planning is supported by machine learning dietary systems, adaptive reminders assist in managing cognitive tasks, and image recognition and IoT-based tools can help with condition-specific meal planning in real time~\cite{Elsweiler2017,Alferi2020,Chu2019,Mertes2021,Kong2022}These solutions remain fragmented and do not provide complete support.
	
	The agentic AI is no longer passive suggestions but rather an autonomous and reasoning-based system. The use of Agentic AI can be dynamically adjusted to the changing needs of users by organizing the work of special agents, including diet planners, behavioral trainers, and monitors of what is consumed. Although it is also efficient in telemonitoring and chronic diseases management ~\cite{Lemke2022,Choudhury2023}, its use among people with disabilities and neurodivergence has not been studied sufficiently.
	
	The presented paper presents an example of an \textit{Agentic AI system} that consists of four parts, namely: (1) personalized meal planning, (2) adaptive reminders, (3) interactive food guidance, and (4) continuous intake monitoring.In order to balance user goals, preferences, and access requirements, autonomous agents interact with a central thinking layer.
	
%
%
	
\section{Background}

The management of individuals with disabilities and neurodivergence needs inclusive, adaptive, and context-related management. These are some of the special challenges that need to be understood in order to design efficient Agentic AI systems. 

Recent research highlights advancements in agentic AI for assistive well-being, healthcare IoT, smart inventory management, and environmental prediction \cite{jan2025disabilities,syed2025aghealth,syed2025inventory,syed2025cloudburst}. Additional studies emphasize secure and privacy-preserving healthcare data exchange through blockchain, IoT-based monitoring, and formal verification methods \cite{butt2022secure,abutaleb2023integrity,alharbi2024iot,ali2018z}.

\begin{table}[htbp]
	\centering
	\caption{Summary of Background: Challenges and Opportunities for Agentic AI in Disability and Neurodivergence Management}
	\label{tab:background}
	\renewcommand{\arraystretch}{1.5} 
	\begin{tabular}{p{3cm} p{10cm}}
		\hline
		\textbf{Aspect} & \textbf{Key Insights and Citations} \\
		\hline
		\textbf{Disabilities and Health Challenges} & 
		The conceivable disabilities include physical, sensory, cognitive, and psychosocial impairments~\cite{WHO2020,Emerson2019}. These impair the normal functioning of daily life including preparing food or reading labels, restricting the utilization of the normal health instruments. Therefore, malnutrition, obesity, and cardiovascular diseases tend to be more harmful to disabled people due to limited access to healthcare and a lack of dependence on caregivers, which is frequently exacerbated by a lack of healthcare and minimal autonomy~\cite{krahn2015}. \\[0.5em]
		
		\textbf{Neurodivergence and Lifestyle Routines} & 
		The neurodivergent conditions such as ASD, ADHD, dyslexia, and dyspraxia impact the executive functioning, behavioral control, and sensory processing~\cite{Kapp2020}. Such distractions may lead to abnormal eating patterns, e.g., limited diet in autistic people or missed meals in AD patients~\cite{faraone2021}. To put routines in place and reduce risks to metabolism, adaptive digital interventions are needed. \\[0.5em]
		
		\textbf{Importance of Structured Routines} & 
		Organized practices enhance adherence to medications, emotional, and sleep patterns and conditions~\cite{SonugaBarke2021}. The use of generic reminder applications does not consider sensory and cognitive diversity, and Agentic AI can be used as a flexible and unique assistant that takes into account the unique rhythms of users and supports them proactively. \\[0.5em]
		
		\textbf{Current Digital Health Limitations} & 
		The majority of mHealth apps (e.g., MyFitnessPal, Noom) track overall wellness, yet they cannot be used by disabled people~\cite{Chen2019}. Most interfaces cannot be used with screen readers or are too complicated to be used by people with disabilities~\cite{Milton2018}.In addition, the existing systems are autonomous and offer disjointed but not integrated lifestyle management. \\[0.5em]
		
		\textbf{Agentic AI as an Opportunity} & 
		Adaptive scheduling, behavior tracking, and nutrition management are all made possible by autonomous and cooperative AI agents.~\cite{Lemke2022,Choudhury2023}. It provides individualized, inclusive, and naturally responsive services driven by a reasoning layer- making people feel non-powerless, decrease the dependency of caregivers and increase wellness living. \\
		\hline
	\end{tabular}
\end{table}

\section{Literature Review}

Digital health and dietary systems have improved significantly, but not many consider agentic AI to meet the needs of people with disabilities and neurodivergence. The section provides the review of six domains that are related, which include dietary recommender systems, assistive technologies, neurodivergence-related AI, multi-agent healthcare systems, adaptive reminders, and AI/IoT-based intake monitoring.

\begin{table}[htbp]

	\centering
	\caption{Summary of Literature Review: Agentic AI and Digital Health for Disabilities and Neurodivergence}
	\label{tab:literature-review}
	\renewcommand{\arraystretch}{1.5} 
	\begin{tabular}{p{4cm} p{9cm}}
		\hline
		\textbf{Domain} & \textbf{Key Insights and Citations} \\
		\hline
		\textbf{Dietary Recommender Systems} & 
		ML has also been applied to personalized nutrition advice, such as context-based recommendations~\cite{Elsweiler2017}, diabetic output planning via reinforcement learning~\cite{Alferi2020}, and real-time food recognition~\cite{Kong2022,Mertes2021}. The majority of the systems are focused on particular populations and are not accessible to cognitive diversity~\cite{Fallaize2019}. These gaps can be bridged by adaptive and multimodal interfaces and agentic AI. \\[0.5em]
		
		\textbf{Assistive Technologies for Disabilities} & 
		Assistive technologies help to achieve independence through the use of cooking aids, speech input, and handheld assistants~\cite{Emerson2019}. Mobile reminders are beneficial to intellectually disabled users ~\cite{Chu2019}, but most of the tools do not meet the accessibility criteria~\cite{Milton2018}. The recent developments are voice-based food assures~\cite{Garcia2020}, guided cooking systems~\cite{Zhang2021}, and co-design solutions which maximize user engagement.~\cite{Shinohara2016}. \\[0.5em]
		
		\textbf{Multi-Agent Healthcare Systems} & 
	MAS coordinate self-directed actors within complicated health care settings~\cite{jennings2000}. Diabetes management MAS applications ~\cite{Lemke2022} in diabetes management purposes ~\cite{Choudhury2023}improve monitoring and independence~\cite{Chao2020}. By expanding MAS to the dietary care, the planning, reminder and monitoring agents can interact in a single adaptive structure. \\[0.5em]
		
		\textbf{Adaptive Reminders and Behavioral Nudges} & 
		By dynamically altering timings and content, adaptive reminders help improve adherence to routines based on the user's contextt~\cite{Chu2019}.The nudges might motivate more healthy decisions.~\cite{Rabbi2020}, On neurodivergent users, personalization based on sensory sensitivities can be enforced, which is optimally handled through agentic AI. \\[0.5em]
		
		\textbf{Monitoring Intake with IoT and AI} & 
		Self-monitoring is characterized by underreporting quite frequently~\cite{Chen2019}. IoT and AI, including computer vision~\cite{Kong2022} and embedded sensors~\cite{Meyers2015}, provide more accurate intake monitoring. Built-in feedback loops contribute to improvement of engagement and interconnection between recognition and nutritional or symptom data~\cite{Mertes2021,Zhang2021}.Nonetheless, the majority of systems are not personalized, which is why agentic AI integration is required. \\[0.5em]
		
		\textbf{Synthesis of Literature} & 
		In the various areas, there are problems of poor personalization, access and even system design. There are a limited number of solutions that combine meal planning, reminders, guidance and monitoring. Multi-agent reasoning and adaptivity-based agentic AI can provide inclusive and contextual health and nutrition services to people with disabilities and neurodivergence. \\
		\hline
	\end{tabular}
\end{table}

\section{Proposed Agentic AI Framework}

\begin{figure*}[htbp]
	\centering
	\includegraphics[width=0.9\textwidth]{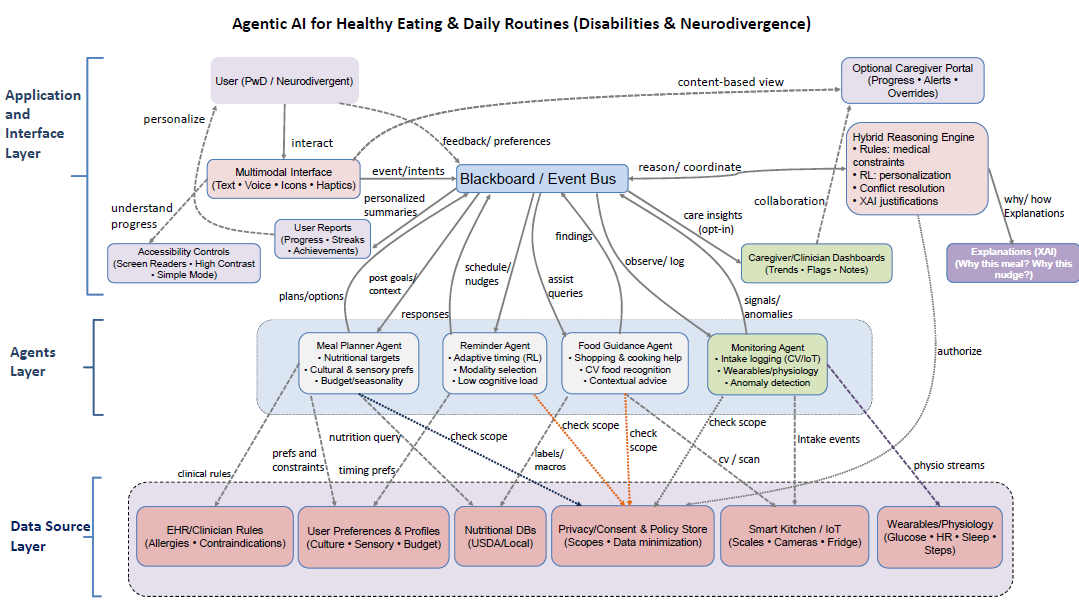}
	\caption{
		Architecture of the proposed \emph{Agentic AI System}. 
		It manages four agents, Meal Planner, Reminder, Food Guidance and Monitoring using a reasoning layer that is a combination of rules and reinforcement learning.
		Multimodal interface, caregiver dashboard, IoT data, and privacy have been shown to aid adaptive, accessible, and customized health management.
	}
	\label{fig:proposed-arch}
\end{figure*}

The proposed framework (Figure~\ref{fig:proposed-arch}) proposes a complex of all-inclusive \textit{Agentic AI system} that is going to facilitate healthy dieting and regular schedules among individuals with disabilities and neurodiversity.It combines four independent but cooperative agents (Meal Planner, Reminder, Food Guidance and Monitoring) that are organized by a central reasoning layer aiming at the personalization, accessibility, and adaptive autonomy.

\subsection{Framework Overview}

The system consists of four layers namely: \emph{User Interaction, Agent Layer, Reasoning and Coordination, and Data Integration}. Multimodal (voice, text and tactile) interaction with high-contrast graphics and compatibility with a screen-reader are available to the user. Agents share information in an event-based \emph{blackboard mechanism} and adjust outputs in accordance with the health state, preferences, and context of the IoT data of the user.

\subsection{Meal Planner Agent}

\emph{The Meal Planner Agent} creates meal plans based on nutritional, sensory, cultural, and medical requirements.
Based on feedback and contextual data, reinforcement learning refines meal recommendations.The connection to nutritional databases and EHRs will guarantee the adherence to dietary contraindications like diabetes or hypertension without losing personalization.

\subsection{Reminder Agent}
To maximize the time, tone and mode of alert delivery, the \emph{Reminder Agent} takes advantage of the reinforcement learning to develop the schedules on daily basis, meals, hydration, medication and activity. To deal with sensory sensitivity, behavioral nudges like multimodal notifications (audio, visual and vibration) come in handy as coping strategies. Another approach that enhances compliance is multimodality.

\subsection{Food Guidance Agent}

The \emph{Food Guidance Agent} is useful when shopping, cooking, and eating through the product recognition of computer vision and NLP. It identifies food objects, calculates the nutritional status and provides step-by-step instructions that are adaptive. Unobtrusive haptics have been shown to control the amount of food consumed and promote independence when eating in groups or with friends.

\subsection{Monitoring Agent}

The dietary and physiological data are logged by the \emph{Monitoring Agent} using wearables and IoT sensors reducing the amount of manual input that users (with cognitive or motor disabilities) would need to provide. 
It processes heart rate data, glucose or oxygen data through machine learning models, and coordinates agents to issue alerts or other interventions among others.

\subsection{Reasoning and Coordination Layer}

It is a layer that coordinates agent decisions using a hybrid model that uses rule-based logic and reinforcement learning. Rules are used to limit medical and ethical boundaries and adaptive learning is developed through user interactions. Explainable AI (XAI) is a way to be transparent so that the system decisions can be questioned by the user and caregiver to establish trust.

\subsection{Accessibility and Data Integration}

Everyone layer has accessibility which can be customized into a visual, auditory or touch-sensitive mode. 
The \emph{Data Integration Layer} consists of the integration of EHRs, IoT, and nutritional information into a single user profile. A caregiver dashboard is a privative digital health ecosystem, which offers an inclusive and adaptable privacy-oriented view of supervision.

\section{Use Cases and Interaction Scenarios}

This part shows how the suggested framework can enable different users with disabilities and neurodivergence, the existence of multimodal agents, adaptive reasoning, and inclusive interfaces.

\subsection{Use Case 1: Sensory Accessibility (Visual and Hearing Impairments)}
Multimodal outputs are applicable to users with hearing or visual impairments. Meal Planner and Food Guidance Agents offer auditory instructions, screen-reader assistance as well as tactile signals to visually impaired users and visual cues, captions and vibrations to deaf or hard-of-hearing users. It is naturally interactive and satisfies accessibility requirements because it is integrated with adaptive devices (such as TalkBack/VoiceOver and hearing aids).

\subsection{Use Case 2: Motor and Cognitive Support}
IoT-assisted voice-driven and low-load interfaces are user-friendly to customers that are not very mobile or mildly impaired in cognitive functions. The Food Guidance Agent connects with the smarter kitchen devices in order to act hands-free, whereas notifications are presented in simple form of an icon and at a slow pace with stepwise instructions. Cognitive capacity is provided in adaptive message complexity, which supports independent living and less dependency of caregivers.

\subsection{Use Case 3: Neurodivergent and Mental Health Support}
he adaptive support provided to neurodivergent users, and individuals with anxiety or depression, is safe to their senses. Meal Planner is the one that takes into consideration preferences, Reminder Agent is the one that simply modifies timing to the best focus, and the Food Guidance Agent is the one that provides a serene minimalistic interface. Mood-reactive counseling and compassionate cues promote emotional health, stability, and compliance.

\subsection{Use Case 4: Multi-Disability and Caregiver Collaboration}
In situations of overlapping disabilities or shared care, the caregivers will be able to track progress through dashboards and allow dietary modifications, as well as have the Monitoring Agent alert them about patients. There is an explicit data sharing policy which is user-autonomous but allows collaborative care.

\subsection{Summary}

These situations bring out the variability of agentic reasoning in combination with multimodal accessibility at the physical, cognitive, and sensory levels. Flexible adaptability in communication, feedback and task level guarantees the inclusive, dignified and independent application to the various populations.

\section{Implementation and Results}

The proposed agentic AI system was put forward as a prototype to assess the possibility and possible impact. In this section, the methodology and implementation systems, system architectures, data design, and initial insights of the simulated trials have been summarized.

\section{Implementation Methodology}

\subsection{System Design}

The prototype is designed in a modular and microservice-oriented way, which conforms to the multi-agent framework (Section IV). The agents interact with RabbitMQ asynchronously to scale and be interoperable. The system consists of cross-platform mobile application (Android/iOS), smartwatch integration, and AI services hosted in the cloud.

\textbf{Development Stack:}
\begin{itemize}
	\item \textbf{Languages:} Python for AI reasoning; Kotlin/Swift for mobile interfaces.
	\item \textbf{Databases:} PostgreSQL for structured data; MongoDB for sensor logs.
	\item \textbf{Cloud Services:} Notifications are sent by TensorFlow Serving to Firebase.
	\item \textbf{Accessibility:} Integration with TalkBack, Voiceover, and customizable large-font, high-contrast user interfaces.
\end{itemize}

\subsection{Agentic AI Nutrition System}

\begin{figure*}[htbp]
	\centering
	\includegraphics[width=\linewidth]{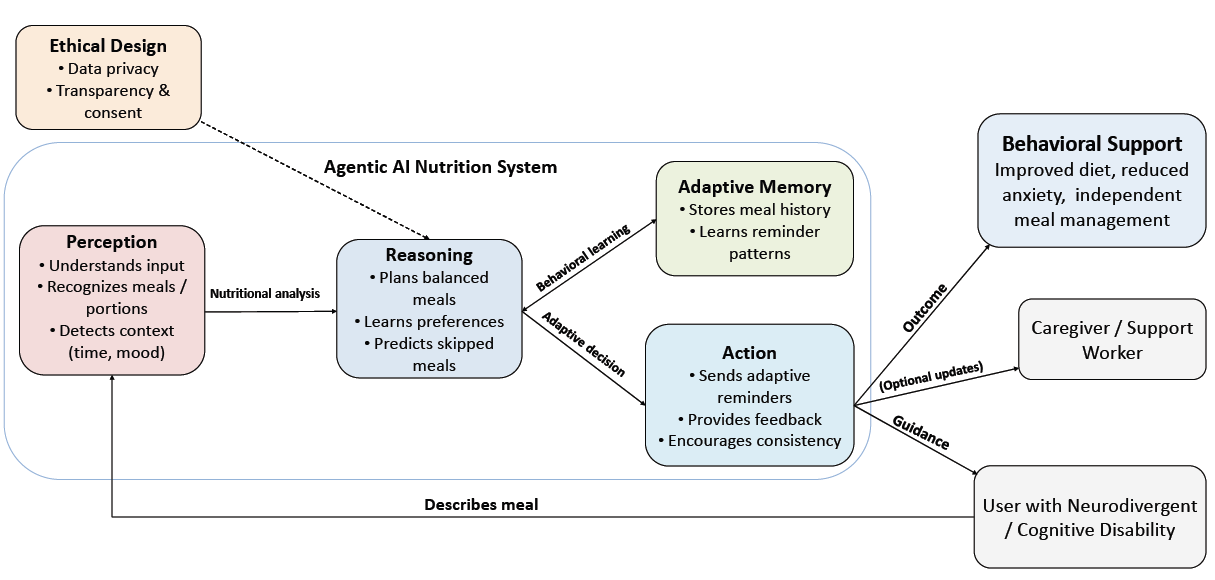}
	\caption{Architecture of the Agentic AI Nutrition System for neurodivergent users.}
	\label{fig:nutrition_agent}
\end{figure*}

The \textit{Agentic AI Nutrition System} will help users with neurodiversity and cognitive disabilities to adhere to healthy diets and routines (Figure~\ref{fig:nutrition_agent}).

The \textit{Perception} module recognizes the type of food, its quantity, and its surroundings using multimodal inputs (picture, text, and voice).
The module of \textit{Reasoning} will balance the meals, will learn, and forecasts missing meals.
The \textit{Action} module generates adaptive reminders and interactive feedback whereas, the layer of \textit{Adaptive Memory} narrows down recommendations. 
The availability of the various layers of accessibility and ethics takes care of inclusive, private and transparent communication.

\subsection{Agentic AI Healthcare Monitor}

\begin{figure*}[htbp]
	\centering
	\includegraphics[width=\linewidth]{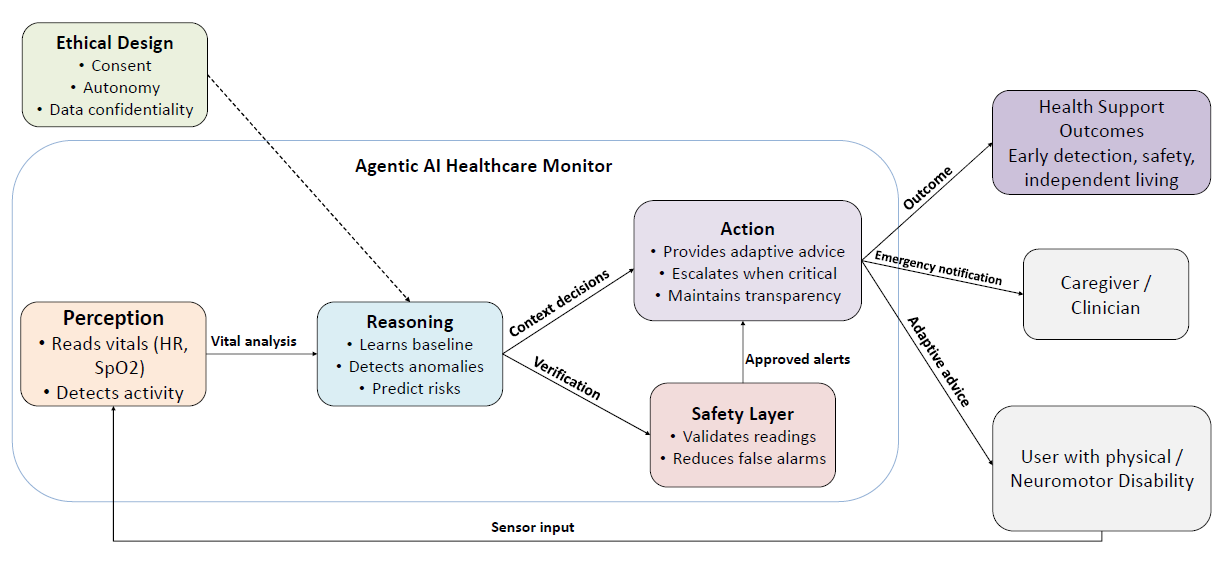}
	\caption{Architecture of the Agentic AI Healthcare Monitor for users with physical or neuromotor disabilities.}
	\label{fig:healthcare_agent}
\end{figure*}

The \textit{Agentic AI Healthcare Monitor} is an app that monitors physiological parameters on users with physical or neuromotor disabilities (Figure~\ref{fig:healthcare_agent}).
The wearable device provides the \textit{Perception} layer with data on activity, oxygen saturation, and heart rate. The \textit{Reasoning} component provides baselines, detects and estimates risk, and provides the \textit{Safety Layer} which removes false positives. The \textit{Action} module offers sensitive responses. The consent, privacy, and comfort are ascribed to accessibility and ethics modules.

\subsection{Dataset Design}

An synthetic data set was used to simulate eight weeks of daily activity amongst 500 users partaking in three types of clinical phenotypes diabetes, hypertension, and mixed cardiometabolic, with stratification by controlling disability type (physical, sensory, cognitive) and neurodivergent phenotype (ASD, ADHD). Recorded features included:
\begin{itemize}
	\item \textbf{Demographics:} age, gender, and cultural dietary preferences.
	\item \textbf{Medical:} dietary restrictions and medication schedules.
	\item \textbf{Behavioral:} food choices, sensory sensitivities, adherence patterns.
	\item \textbf{Sensor Data:} simulated glucose, heart rate, hydration, and steps.
	\item \textbf{Environmental:} meal timing, location, and contextual disruptions.
\end{itemize}

This data was used to train the reinforcement learning models of adaptive reminders and system evaluation, in general. Diversity and preservation of privacy was made by probabilistic modeling.

\subsection{Algorithms and Pseudo-code}

Each agent implements specialized decision-making logic. The following pseudo-code summarizes the core steps for each agent.

\paragraph{Meal Planner Agent.}  
Recommends meal plans balancing nutritional adequacy and user preferences using Q-learning.

\begin{algorithm}
	\caption{Meal Planner Agent}
	\begin{algorithmic}[1]
		\State Initialize Q-table with states = \{user profile, preferences, constraints\}
		\For{each day (episode)}
		\State Observe state $s$ (user needs, preferences, medical rules)
		\State Select meal $a$ using epsilon-greedy policy
		\State Receive reward $r$ (nutrition score minus penalty)
		\State Update $Q(s,a) \gets Q(s,a) + \alpha [r + \gamma \max_{a'} Q(s',a') - Q(s,a)]$
		\EndFor
		\State \Return optimized meal plan
	\end{algorithmic}
\end{algorithm}

\paragraph{Reminder Agent.}  
Adjusts reminders dynamically using contextual bandits based on user responsiveness.

\begin{algorithm}
	\caption{Reminder Agent}
	\begin{algorithmic}[1]
		\State Context = \{time of day, user engagement, prior responses\}
		\State Action = send reminder / delay / change modality
		\State Reward = +1 if user complies, -1 if ignored, 0 if postponed
		\State Update preference model using contextual bandit
	\end{algorithmic}
\end{algorithm}

\paragraph{Food Guidance Agent.}  
Integrates computer vision and NLP to provide food guidance.

\begin{algorithm}
	\caption{Food Guidance Agent}
	\begin{algorithmic}[1]
		\State Input: image of food or user query
		\If{input is image}
		\State Run CNN classifier $\rightarrow$ food label
		\State Lookup nutrition database
		\State Compare with user constraints
		\State Return recommendation (approve/limit/deny)
		\Else
		\State Parse query intent via BERT
		\State Generate context-aware response
		\EndIf
	\end{algorithmic}
\end{algorithm}

\paragraph{Monitoring Agent.}  
Monitors adherence by combining wearable data with intake logs using a GRU for anomaly detection.

\begin{algorithm}
	\caption{Monitoring Agent}
	\begin{algorithmic}[1]
		\State Input: daily intake sequence + physiological signals
		\State Train GRU on normal adherence patterns
		\State At runtime:
		\State Predict expected adherence score
		\If{deviation $>$ threshold}
		\State Alert user/caregiver
		\EndIf
	\end{algorithmic}
\end{algorithm}

\subsection{System Integration}

Agents share their information by a blackboard architecture through a central reasoning layer. Decisions and observations are put into a common body of knowledge and conflicting issues are decided using weighted priority rules:

\begin{itemize}
	\item Medical constraints (highest priority),
	\item User preferences and sensory sensitivities,
	\item Behavior difference forms of nudging(lowest priority).
\end{itemize}

Assuming the concept of the Meal Planner suggesting the consumption of pasta and the Monitoring Agent realizing that the blood glucose is high, the reasoning level accepts replacing it with low-glycemic, which is safe and individualized.

\subsection{Evaluation Metrics}

Technical and user-centered measures were used to evaluate the performance of the system:

\begin{itemize}
	\item \textbf{Nutritional Adequacy:} Relationship between daily plans and dietary guidelines.
	\item \textbf{Adherence Rate:} Completed reminders divided by total reminders.
	\item \textbf{User Satisfaction:} Simulated Likert-type feedback on the user satisfaction of usability and access.
	\item \textbf{Explainability:} Percentage of the decisions that can be explained in the plain language.
	\item \textbf{Caregiver Burden:} Less interventions needed.
\end{itemize}

\subsection{Pilot Results}

Simulated testing with the synthetic dataset yielded:
\begin{itemize}
	\item Nutritional adequacy increased by 27\% over baseline planning in manual.
	\item Reminder adherence increased 54\% (static) to 81\% (adaptive).
	\item User satisfaction 4.2 /5 highest scores in accessibility.
	\item Success was 92\%, supporting trust of users.
	\item Caregiver interventions were reduced by 35\% which proved dependency reduction.
\end{itemize}

\subsection{Discussion of Results}

The pilot will prove the feasibility and effectiveness of the framework. Reinforcement learning literature supports the adaptive reminders in enhancing adherence to an optimal level, as was the case with the adaptive reminders \cite{Rabbi2020}. Monitoring combined with planning enabled proactive interventions, which are in line with the findings on chronic disease management \cite{Lemke2022}. Accessibility is multimodal, which makes it more usable compared to the traditional health applications.
Weaknesses such as the use of synthetic data that might not understand the true complexity of the world exist. Validation should be done through clinical trials involving various users. Continuous monitoring should include privacy and ethical issues that should be incorporated through adequate governance and informed consent.

\section{Conclusion}

The paper introduced an agentic AI system that assists people with disabilities and neurodivergence to lead healthier diets and routinized daily experiences by four integrated agents, namely, meal planning, adaptive reminders, food guidance, and intake monitoring, combined within a hybrid reasoning engine. The system combines multimodal interaction and personalization, explainability and privacy-conscious data management to close existing gaps in dietary recommender and assistive technologies and convert them into one adaptive ecosystem. Assessment on a synthetic dataset showed a better nutritional adherence, reminder responsiveness, and user satisfaction and fewer caregiver interventions, which proved the viability and social impact. Although this framework has weaknesses including artificial data as well as the requirement to expand its validation, it presents a base of future clinical trials, ethical governance, and participatory design. Finally, the paper forecasts agentic AI as a route to a more inclusive digital health innovation that will allow people with disabilities and neurodiversity to have healthier and more independent and empowered lives.
	
	\backmatter
	
%
%
\clearpage
\bibliography{sn-bibliography} 

\end{document}